\documentclass[conference]{IEEEtran}
\IEEEoverridecommandlockouts
\usepackage{cite}
\usepackage{amsmath,amssymb,amsfonts}
\usepackage{algorithmic}
\usepackage{graphicx}
\usepackage{textcomp}
\usepackage{xcolor}
\usepackage{hyperref}
\usepackage{fancyhdr}

\def\BibTeX{{\rm B\kern-.05em{\sc i\kern-.025em b}\kern-.08em
    T\kern-.1667em\lower.7ex\hbox{E}\kern-.125emX}}
\begin{document}

\title{Unsupervised and Interpretable Domain Adaptation to Rapidly Filter Tweets for Emergency Services
}

\author{\IEEEauthorblockN{Jitin Krishnan}
\IEEEauthorblockA{\textit{Department of Computer Science} \\
\textit{George Mason University}\\
Fairfax, VA, USA\\
jkrishn2@gmu.edu}
\and
\IEEEauthorblockN{Hemant Purohit}
\IEEEauthorblockA{\textit{Department of Information } \\
\textit{Sciences \& Technology}\\
\textit{George Mason University}\\
Fairfax, VA, USA\\
hpurohit@gmu.edu}
\and
\IEEEauthorblockN{Huzefa Rangwala}
\IEEEauthorblockA{\textit{Department of Computer Science} \\
\textit{George Mason University}\\
Fairfax, VA, USA\\
rangwala@gmu.edu}
}

\maketitle
%%%%%%%%%%%%%%%%%%%%%%%%%%%%
\thispagestyle{fancy}
%%%%%%%%%%%%%%%%%%%%%%%%%%%%

\begin{abstract}
% Motivation
During the onset of a natural or man-made crisis event, public often share relevant information for emergency services on social web platforms such as Twitter. 
%through web platforms, especially social web. 
However, filtering such relevant data %information 
in real-time at scale using %machine learning 
%automated techniques 
social media mining 
is challenging due to the short noisy text, sparse availability of relevant data, and also, practical limitations in collecting large labeled data during an ongoing event. %crisis. % for such data. for an ongoing event. In such scenarios, 
In this paper, we hypothesize that unsupervised domain adaptation through multi-task learning can be a useful framework %that can make use of data from previous crisis situations 
to leverage data from past crisis events for training efficient information filtering models during the sudden onset of a new crisis. 
We present a novel method to classify relevant social posts during an ongoing crisis without seeing any new data from this event (fully unsupervised domain adaptation). %
%In this paper, we propose a novel method to classify relevant social media posts during a crisis event %low-resource data collected from the web during crises; focusing specifically on short bursts of Twitter post (tweet) streams.
% 
%Additionally, %we focus on a crucial component that is 
%our method addresses a crucial but missing component from current research for social media analytics during crises: \textit{model interpretability}.  
% Methods 
% 
%Toward the goal of domain adaptation and interpretability, 
Specifically, %we first identify a standard single-task attention-based neural network architecture  %\cite{jitin} 
%and empirically compare it to the state-of-the-art methods using the standard benchmark %multi-domain dataset of Amazon reviews. Using this baseline as the building block, we 
%and then 
we construct a customized multi-task architecture with a multi-domain discriminator for crisis analytics: \textit{multi-task domain adversarial attention network} (MT-DAAN). This model consists of dedicated attention layers for each task to provide \textit{model interpretability}; critical for real-word applications. 
% Results 
As deep networks struggle with sparse datasets, we show that this can be improved by sharing a base layer for multi-task learning and domain adversarial training. 
The framework is validated with the public datasets of TREC incident streams that %we classify 
 provide labeled Twitter posts (tweets) with relevant %mutually inclusive 
 classes (\textit{Priority}, \textit{Factoid}, \textit{Sentiment}) %and \textit{Irrelevant}) 
 across 10 different crisis events such as floods and earthquakes. %, and Shootings.
Evaluation of domain adaptation for crisis events is performed by choosing one target event as the test set and training on the rest. 
Our results show that the multi-task model outperformed its single-task counterpart. For the qualitative evaluation of interpretability, we show that the attention layer can be used as a guide to explain the model predictions and empower emergency services for exploring accountability of the model, by showcasing the words in a tweet that are deemed important in the classification process. 
Finally, we show a practical implication of our work by providing a use-case for the COVID-19 pandemic. 
\end{abstract}


\begin{figure}[h]
  \centering
    \includegraphics[width=8.3cm]{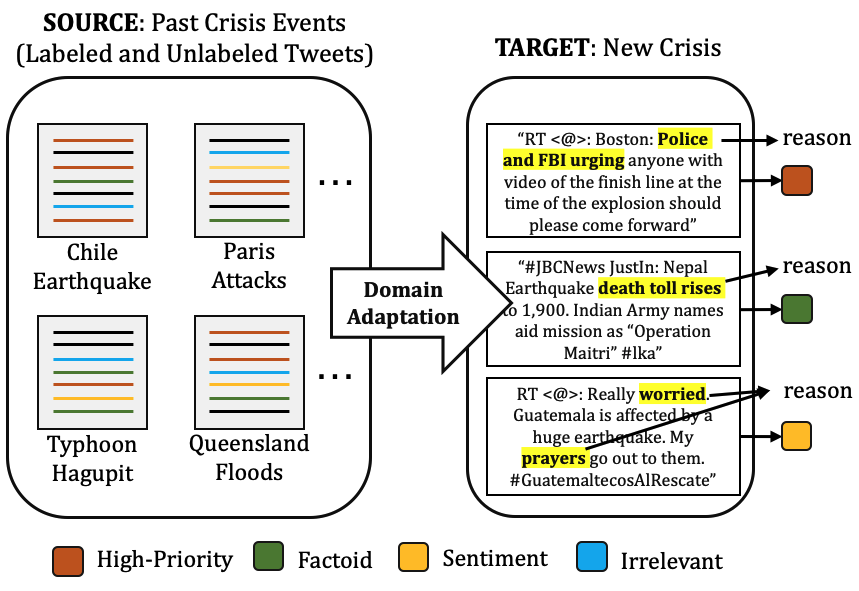}
 \caption{\textbf{Problem Statement}: Interpretably predict labels for tweets collected during an ongoing crisis using only the past crisis data, given a) unavailability of labeled data in the ongoing event, and b) need for interpretability of machine reasoning behind data filtering for emergency managers.}\label{fig:mtl} 
\end{figure}

\section{Introduction}
During the sudden onset of a crisis situation, social media platforms such as Twitter provide valuable information to aid crisis response organizations in gaining real-time situational awareness~\cite{castillo2016big}. Effective analysis of important information such as affected individuals, infrastructure damage, medical emergencies, or food and shelter needs can help emergency responders make time-critical decisions and allocate resources in the most effective manner \cite{imran2016twitter,li2018disaster,vieweg2014integrating}. 

Several machine learning systems have been deployed to help towards this humanitarian goal of converting real-time social media streams into actionable knowledge. Classification being the most common task, researchers have designed models \cite{nguyen2016rapid,alam2018domain,mazloom2019hybrid,li2018disaster} that classify tweets into various crisis-dependent categories such as priority, affected individuals, type of damage, type of assistance needed, usefulness of the tweet, etc. Social media streams contain short, informal, and abbreviated content; with potential linguistic errors and sometimes contextually ambiguous. These inherently challenging properties of tweets make their classification task and formulation less trivial when compared to traditional text classification tasks.

In this paper, we address two practically important and underdeveloped aspects of current research in social media mining for crisis analytics to classify relevant social web posts: \textbf{a)} a fully unsupervised domain adaptation, and \textbf{b)} interpretability of predictions. A fully unsupervised domain adaptation uses \textbf{no data} from the ongoing crisis to train the model. Nguyen et al., 2016 \cite{nguyen2016rapid} showed that their convolutional neural network (CNN) model does not require feature engineering and performed better than the state-of-the-art methods; one of their models being completely unsupervised \cite{nguyen2016rapid}. Similarly, Alam et al., 2018 ~\cite{alam2018domain} designed a CNN architecture with adversarial training on graph embeddings, but utilizing unlabeled target data. 
Our goal is to construct an unsupervised model that does not require any unlabeled target data with the capability of being interpretable. We specifically address the problem of data sparsity and limited labels by designing a multi-task classification model with domain adversarial training; which, to the best of our knowledge, is not explored in social media mining for crisis analytics. Another crucial component of our model is interpretability. In prior works, when a top performing model produces an accuracy of $78\%$, for instance, it is unclear how trustworthy it is and what features are deemed important in the model's decision-making process. An interpretable model like ours can present with a convincing evidence of which words the classifier deems important when making a certain prediction, and helps ensure reliability for domain users, e.g., emergency managers. %This also brings additional benefits of using them in downstream tasks such as knowledge graph construction. % and creating nodes for social network analysis.

\textbf{Contributions:} 
%\subsection{Contributions}
\textbf{a)} To address the problems of data sparsity and limited labels, we construct a customized multi-task learning architecture (MT-DAAN) to filter tweets for crisis analytics by training four different classification tasks (\textit{c.f.} examples in Fig.~\ref{fig:attn_viz}) across ten different crisis events under domain shift. This multi-task domain adversarial model consists of dedicated attention layers for each task for interpretability and a domain classifier branch to promote the model to be domain-agnostic. \textbf{b)} We demonstrate that the attention layers provide interpretability for the predictions made by the classifiers; with the goal to aid emergency services in a more meaningful way. \textbf{c)} We %identify a standard single-task attention-based neural network architecture and 
empirically validate the performance of the underlying single-task attention-based neural network architecture by comparing it to the state-of-the-art methods, for improving generalizability and interpretability for domain adaptation in unsupervised tweet classification tasks in general. \textbf{d)} Additionally, through experiments, we show that deep networks struggle with small datasets, and that this can be improved by sharing the base layer for multi-task learning and domain adversarial training.

%The rest of the paper is organized as follows. Section 2 presents related works to our methodology: \textit{Domain Adaptation}, \textit{Mult-Task Learning}, \textit{Attention}, and \textit{Interpretability}. Section 3 presents the methodology: MT-DAAN. Section 4 presents the datasets and experimental set up. Finally, the results are discussed in section 5, future directions in section 7, and conclusion in section 8. Section 6 provides a use-case for the ongoing COVID-19 pandemic. 

\section{Related Work and Background}
%\quad \textbf{Attention:} Attention mechanism

%\textbf{Domain Adaptation:} 
\subsection{Domain Adaptation}
Domain Adaptation in text classification tasks has a long line of fruitful research \cite{blitzer2006domain,pan,sda} that try to minimize the difference between the domains so that a model trained solely on one domain is generalizable to unseen test data from a completely different domain. With the introduction of Domain-Adversarial training of Neural Networks (DANN) \cite{ganin2016domain}, many state-of-the-art models now utilize unlabeled target data to train classifiers that are indiscriminate toward different domains. The speciality of this architecture is that it consists of an extra branch, which performs domain classification using unlabeled data from different domains. Thus, both task and domain classifiers share some bottom layers but have separate layers towards the top. A negative gradient (gradient reversal) from the domain classifier branch is back-propagated to promote the features at the lower layers of the network incapable of discriminating the domains. Recent works such as Adversarial Memory Network (AMN) \cite{amn} utilizes attention, in addition to DANN, to bring interpretability to capture the pivots for sentiment classification. Hierarchical Attention Network (HATN) \cite{li2018hierarchical} expands upon AMN by first extracting pivots and then jointly training networks for both pivots and non-pivots. 

For filtering social web data for crisis analytics, these models do not suffice and need customized expansions due to the following reasons: \textbf{a)} Collecting and using large unlabeled target data from the new/ongoing crisis event may not be practically viable, %during a new crisis event;
thus, we aim for a fully unsupervised modeling. %\textbf{b)} 
\textbf{b)} Having access to unlabeled data from multiple crisis events can alleviate the above problem to an extent by using it to train the domain classifier branch to push the model to be domain independent. %and %\textbf{c)} 
\textbf{c)} Due to the low-resource nature of the dataset, binary classifiers may miss important lower level features that can be potentially improved by a multi-task model that shares the lower layers of the network for all the tasks. This is also evident from our results in \textbf{Table \ref{table:benchmark}} and \textbf{\ref{table:simple_benchmark}}, which show that deep models that perform much better than simple models on Amazon reviews do not significantly outperform them on TREC tweet dataset for crises.

%\textbf{Multi-Task Learning:} 
\subsection{Multi-Task Learning}
Multi-Task Learning (MTL) solves multiple tasks at the same time with a goal to improve the overall generalization capability of the model \cite{caruana1997multitask}. Within the context of Deep Learning, MTL is performed by sharing (or constraining) lower level layers and using dedicated upper level layers for various tasks. A rich overview of MTL in Deep Neural Networks is presented by Ruder (2017) \cite{ruder2017overview}. MTL has been a successful strategy over the past few years for many research explorations such as relationship networks \cite{long2015learning} in computer vision and Sluice networks \cite{ruder122017sluice} in natural language processing. Similar problems in domain adaptation of semantic classification and information retrieval were addressed by jointly learning to leverage large amounts of cross-task data \cite{liu2015representation}. In low resource datasets such as for crises, the chance of overfitting is very high. %For this reason,
Thus, it seems intuitively better for the model to find a shared representation  %that captures all the 
capturing different tasks and not just one, such that feature commonalities across %different 
tasks can be exploited.  

\subsection{Attention Mechanism}
Attention mechanism \cite{bahdanau2014neural}, originally designed for machine translation problems, has become one of the most successful and widely used methods in deep learning that can look at a part of a sentence at a time like humans. This is particularly useful because of its ability to construct a context vector by weighing on the entire input sequence unlike previous sequence-to-sequence models \cite{sutskever2014sequence} that used only the last hidden state of the encoder network (typically BiLSTM \cite{schuster1997bidirectional}, LSTM \cite{hochreiter1997long}, or GRU \cite{chung2015gated}). For example, in a sentence, the context vector is a dot product of the word activations and weights associated with each word; thus leading to an improved contextual memorization, especially for long sentences. Our method incorporates such attention mechanisms to enhance interpretability of the classifier.

%\subsection{Interpretability}
%With more and more machine learning systems being adopted by diverse application domains, transparency in decision-making inevitably becomes an essential criteria, especially in high-risk scenarios \cite{gunning2017explainable} where trust is of utmost importance.  With deep neural networks, including natural language systems, shown to be easily fooled \cite{nguyen2015deep, jia2017adversarial}, there has been many promising ideas that empower machine learning systems with the ability to explain their predictions \cite{ribeiro2016should,kingma2013auto,chen2016infogan}. \citeauthor{gilpin2018explaining} \cite{gilpin2018explaining} presents a survey of interpretability in machine learning, which provides a taxonomy of research that addresses various aspects of this problem. Similar to the work by \citeauthor{ross2017right} \cite{ross2017right}, we employ an attention-based approach to evaluate model interpretability applied to the crisis-domain.

\section{Methodology}

\subsection{Problem Statement: Unsupervised Domain Adaptation for Crisis Tweet Classification} % Definition \& Notations}

Using notations in \textbf{Table \ref{tab:notations}}, consider a set $C$ of all crisis events such as \textit{Guatemala Earthquake} or \textit{Typhoon Yolanda}. The task of unsupervised domain adaptation for crisis analytics is to train a classifier for a specific target crisis ($c_t$) using labeled ($L_{C-c_t}$) and unlabeled ($U_{C-c_t}$) data from all other crises; where $C-c_t$ denotes the set of all crisis events minus the target crisis. We assume that \textbf{no} data record from the target crisis is available for training. Following the traditional domain adaptation terminology, $X_s$ = $L_{C-c_t}$ represents the labeled data from the source domain $S$ and $Y_s$ = $y_{C-c_t}$ represents the ground truth labels on which the classifier is trained. And, $X_t$ = $L_{c_t}$ represents the labeled data from the target domain $T$ and $Y_t$ = $y_{c_t}$ represents the ground truth labels; both of which are only used for testing the classifier. $X_d$ = $U_{C-c_t}$ represents the unlabeled data from different domains minus the target. To summarize:\\
\textit{\textbf{Input:}} $X_s$, $Y_s$, $X_d$  \\
\textit{\textbf{Output:}} $Y_{t}^{pred}$ $\gets$ $predict(X_t)$ \\

\begin{table}[!htbp]
%\footnotesize
\scriptsize
\begin{center}
 \begin{tabular}{||p{1.2cm}| p{5.9cm} ||} 
 \hline
 \textbf{Notation} & \textbf{Definition} \\ [0.5ex] 
 \hline\hline
 $C$ & Set of all crisis events $\{c_1, c_2, ...\}$ \\
 \hline
 $L_{c_k}$ & Set of labeled data from the event $c_k$ \\
 \hline
 $y_{c_k}$ & Set of ground truth labels for $L_{c_k}$. \\
 \hline
 $m$ & Number of tasks (Number of bits in each label)  \\
 \hline
 $U_{c_k}$  & Set of unlabeled data from the event $c_k$  \\
 \hline
 \hline
  $T_x$  & Number of words in a sentence  \\
 \hline
  $x^{<k>}$  & $k$-th word of a sentence \\
 \hline
  $\alpha^{<k>}$  & attention from $k$-th word \\
 \hline
  $a^{<k>}$  & BiLSTM activation from $k$-th word \\
 \hline
 \end{tabular}
\end{center}
\caption{Notations}\label{tab:notations}
\end{table}

\subsection{Overview}

In the following sections, we describe three models: Single-Task Attention Network (ST), Single-Task Domain Adversarial Attention Network (ST-DAAN), and Multi-Task Domain Adversarial Attention Network (MT-DAAN). ST is the model we adopt from %Krishnan et al. (2020) 
~\cite{jitin} to build the single-task attention based baseline. ST-DAAN is constructed on top of ST to make the model domain agnostic by performing adversarial training using gradient reversal. Finally, MT-DAAN is constructed on top of ST-DAAN with dedicated attention layers for each task on a shared BiLSTM layer. This is shown in \textbf{Figure \ref{fig:mtl}}.

\begin{figure}[h]
  \centering
    \includegraphics[width=8.5cm]{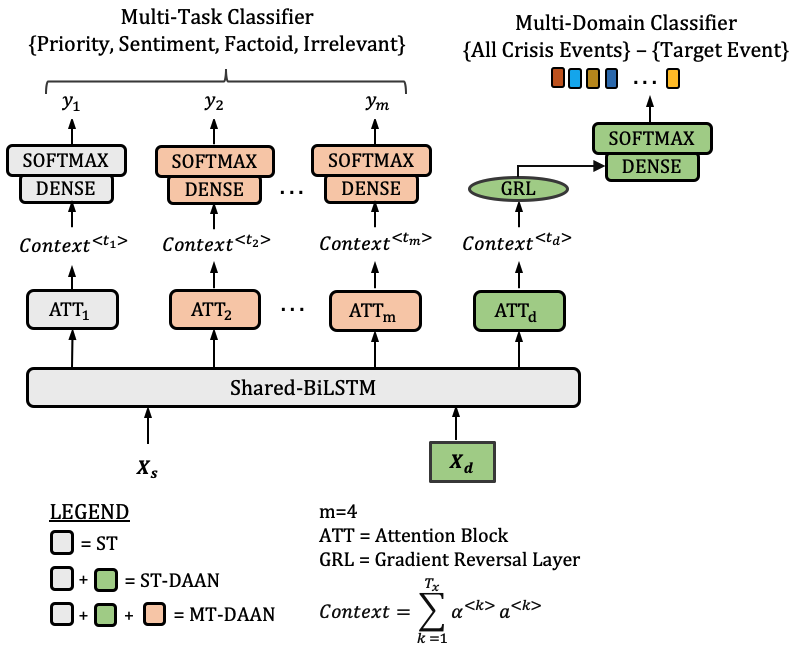}
 \caption{Fully Unsupervised Domain Adaptation Set-up for Multi-Task Crisis Tweet Classification.}\label{fig:mtl}
\end{figure}

\subsection{Single-Task Attention Network (ST)}

We first describe the single-task attention network \cite{jitin} on top of which we build our models. This model aligns with our goals of interpretability and unsupervised domain adaptation. This BiLSTM based model with Attention gives us three main advantages: 
\begin{enumerate}
    \item Unlike several existing domain adaptation methods that use unlabeled target data to train the domain adversarial component via gradient reversal, this method is a fully unsupervised baseline which also can be customized for multi-task learning.
    \item The method uses attention mechanism which in turn weighs each word in a sentence based on its importance. This can be directly utilized for interpretability. 
    \item The method also runs much faster (only a few minutes), i.e. highly useful in crisis times, as compared to the top performing semi-supervised models such as HATN \cite{li2018hierarchical} (hours). 
\end{enumerate}

This model \cite{jitin} consists of a BiLSTM layer which produces $T_x$ activations, each corresponding to a word in the sentence. These activations are passed through \textit{dense} and \textit{softmax} layers and are combined by dot product to produce the context vector $\sum_{k=1}^{T_x} \alpha^{<k>} a^{<k>}$, where $a^{<k>}$ is the BiLSTM activation from $k$-th word and $\alpha^{<k>}$ is the attention weight of $k$-th word. Sentences with words greater than $T_x$ are stripped and those with words lower than $T_x$ are padded. This single-task ($m=1$) attention network is the building block with which rest of the following models are constructed. The single-task binary cross entropy loss function is shown below.
\begin{equation}
\footnotesize
%\scriptsize
L_{T} = -  \frac{1}{N} \sum_{i=1}^{N} [y_i \log \hat{y_i} +  (1-y_i) \log (1-\hat{y_i})] \label{eq:1}
\end{equation}

where $T$ represents the task, $y$ is the true label, and $\hat{y}$ is the predicted label.

\subsection{Single-Task Domain Adversarial Attention Network\\ (ST-DAAN)}

To study the specific contribution of domain adversarial training, we construct a secondary baseline over the ST architecture by constructing an additional branch with gradient reversal layer which is represented by the green blocks in \textbf{Figure \ref{fig:mtl}}. This is a single-task binary classifier with $m=1$. Domain Adversarial Training of Neural Networks (DANN) \cite{ganin2016domain} was introduced with a goal to confuse the classifier by back-propagating a negative gradient from a separate domain classifier branch (right-most branch, as shown in \textbf{Figure \ref{fig:mtl}}). This makes the classifier agnostic to difference in domains. This back-propagation is implemented using a \textit{gradient reversal layer} \cite{ganin2016domain} which does nothing during the forward pass but pushes a negative gradient ($- \lambda \frac{\partial L_d}{ \partial \theta_f}$) during the backward (gradient update) pass. $L_d$ is the domain classification loss, $\lambda$ is the strength of the reversal, and $f$ represents the lower level layers or features over which the negative gradient update is performed. In our architecture, the goal is to make the BiLSTM layer indiscriminate towards various crisis domains such that the multi-task classification does not depend on the domain from which the tweet/sentence is coming from. The ST-DAAN loss function is shown below.

\begin{equation}
\footnotesize
L_{T}' = L_{T} +  w_{d} L_{d} \label{eq:2}
\end{equation}

where $w_d$ is the domain adversarial loss weight. $L_{d}$ represents the categorical cross entropy loss for multi-domain discriminator shown below.

\begin{equation}
\footnotesize
L_{d} = -  \frac{1}{N} \sum_{i=1}^{N} \sum_{j=1}^{|C-c_t|} [y_{ij} \log \hat{y_{ij}}] \label{eq:3}
\end{equation}

where $C-c_t$ is the set of all crisis events without the target event.

\subsection{Multi-Task Domain Adversarial Attention Network \\(MT-DAAN)}
%\subsection{Baseline BiLSTM Attention Network \cite{jitin}}

\iffalse
\begin{figure}[h]
  \centering
    \includegraphics[width=5.5cm]{attnagg.png}
 \caption{Multi-Head Attention Aggregation Methods: \textbf{a)} Sum, \textbf{b)} Max, \textbf{c)} Mean, \textbf{d)} Concatenation. $\vec{C} =$ context vector.} \label{fig:attnagg}
\end{figure}
\fi

%\subsection{Multi-Task Domain Adversarial Attention Network (MT-DAAN)}
Building on top of ST-DAAN, we construct MT-DAAN, which is intended to classify problems with multiple tasks or labels. For each task, a dedicated attention layer is allocated from which it predicts binary labels. The BiLSTM layer remains exactly the same as in the single-task model but multiple attention blocks are added for each task along with a domain classifier. In the architecture decision process, we first investigated a multi-label classifier where all layers are shared with the final softmax layer making multi-label predictions. In low resource settings, constructing a multi-label classifier using a shared architecture is challenging for two reasons: \textbf{a)} jointly balancing positive and negative samples across all classes is not trivial and potentially challenging to make it extensible when new classes need to be added, and \textbf{b)} attention layer may not always produce class-specific insights as the weights are assigned to train for the combination of labels. On the other hand, in the multi-task architecture with separate attention layers, it is easy to add more classes. If some classes require more training, it is trivial to further tune a model specific to that class. More importantly, $context^{<t_j>}$ vector for $j$-th task identifies the influential words from each sentence for that specific task. The complete architecture is shown in \textbf{Figure \ref{fig:mtl}}. MT-DAAN loss function is shown below:

\begin{equation}
\footnotesize
L_{MT-DAAN} = \sum_{k=1}^{m} (w_k L_{T_k}) +  w_{d} L_{d} \label{eq:4}
\end{equation}

where $m$ is the number of tasks, $w_k$ is the loss weight and $L_{T_k}$ is the loss term for each task, $w_d$ is the domain adversarial loss weight, and $L_{d}$ is the domain adversarial loss term.

\subsection{Model Interpretability}

The output ($\alpha$) of the attention layer ($ATT$) of each task, is a $T_x$-dimensional vector; $T_x$ being the number of words in the sentence. The context vector ($\sum_{k=1}^{T_x} \alpha^{<k>} a^{<k>}$) is the product of these attention weights and the $T_x$-dimensional activation ($a$) from the $BiLSTM$ layer. $\alpha$ essentially weighs how much each word in the sentence contributes to the classification result. Thus, $\alpha$ is the component that is evaluated for model interpretability.

\section{DATASETS}

\begin{table}
\scriptsize
%\footnotesize
\begin{center}
 %\begin{tabular}{|p{4.1cm}| p{0.9cm} | p{0.9cm}  p{0.95cm} | p{0.9cm}  p{0.9cm} p{1.2cm}  p{1.2cm} | } 
 \begin{tabular}{|p{2.95cm}| p{0.6cm} | p{0.4cm}  p{0.57cm} | p{0.2cm}  p{0.2cm} p{0.2cm} p{0.2cm} | } 
 \hline
 CRISIS EVENTS & Total Tweets & Vocab & Avg \#words & P & F & S & I \\ 
 \hline
 2012 Guatemala Earthquake & 154 & 422 & 18.74 & 104 & 108 & 12 & 15 \\
 2013 Typhoon Yolanda & 564 & 1746 & 19.47 & 249 & 46 & 119 & 51 \\
 2013 Australia Bushfire & 677 & 2102 & 20.21 & 152 & 213 & 167 & 36 \\
 2013 Boston Bombings & 535 & 1755 & 19.30 & 147 & 28 & 234 & 198 \\
 2013 Queensland Floods & 713 & 2301 & 19.08 & 293 & 54 & 173 & 215 \\
 2014 Chile Earthquake & 311 & 919 & 16.54 & 48 & 26 & 50 & 10 \\
 2014 Typhoon Hagupit & 1470 & 2893 & 15.36 & 469 & 375 & 276 & 101 \\
 2015 Nepal Earthquake & 2048 & 4026 & 13.77 & 1067 & 377 & 741 & 133 \\
 2015 Paris Attacks & 2066 & 4152 & 18.62 & 306 & 183 & 782 & 429 \\
 2018 Florida School Shooting & 1118 & 2940 & 21.40 & 329 & 64 & 206 & 70\\
 \hline
 \end{tabular}
\end{center}
  \caption{TREC Dataset Statistics; Showing the number of positive samples for each of the 4 classes. $P$=Priority, $F$=Factoid, $S$=Sentiment, and $I$=Irrelevant.}\label{table:tweet_stat}
\end{table}

\subsection{TREC Dataset}

TREC-IS\footnote{\url{http://dcs.gla.ac.uk/~richardm/TREC_IS/}} (Text Retrieval Conference - Incident Streams) is a program that encourages research in information retrieval from social media posts with the goal to improve the state-of-the-art social media based crisis analytics solutions. We use the dataset from 2018 track proposal. Statistics of this curated dataset of Twitter downloaded from TREC is shown in \textbf{Table \ref{table:tweet_stat}}. The original dataset consisted of 15 crisis events. However, due to very low data, we trimmed the events and tasks such that there are at least 10 positive samples for each task.

The four tasks used in our experiments are shown below:
\begin{enumerate}
    \item \textit{Priority}: Different priority levels are assigned for each tweet: \textit{low}, \textit{medium}, \textit{high}, \textit{critical}. We convert this into a binary classification problem where $low = 0$ and $\{medium$, $high$, $critical\} = 1$.
    \item \textit{Factoid}: `Factoid' is a categorical label that represents if a tweet is stating a fact. Eg: `\textit{death toll rises ..}.'
    \item \textit{Sentiment}: `Sentiment' is a categorical label that represents if a tweet represents a sentiment. Eg: '\textit{Worried.. Thoughts and prayers}.'
    \item \textit{Irrelevant}: `Irrelevant' is a categorical label for tweets that do not provide any relevant information.
\end{enumerate}

\subsection{Amazon Reviews Dataset }  %(for multitasking)

The standard benchmark dataset\footnote{\url{http://www.cs.jhu.edu/~mdredze/datasets/sentiment/}} of Amazon reviews \cite{blitzer2007biographies} is widely used for cross-domain sentiment analysis. We chose four domains: Books (B), Kitchen (K), DVD (D), and Electronics (E). The raw data\footnote{\url{https://github.com/hsqmlzno1/HATN/tree/master/raw_data}}, a part of Blitzer's original raw dataset, used in this work is from HATN \cite{li2018hierarchical}. This dataset consists of $3000$ positive and $3000$ negative samples for each of the $4$ domains. This dataset is used for two purposes: 1) to validate the performance of the state-of-the-art methods including the single-task baseline and 2) to compare and contrast the performance of deep models when trained with rich versus sparse datasets.

%This dataset is used to compare and contrast the performance of deep models when trained with rich versus sparse datasets. 

\iffalse
\begin{table}
%\scriptsize
\footnotesize
\begin{center}
 \caption{Implementation Details}\label{table:implementation}
 \begin{tabular}{||p{2.8cm}| p{4.4cm} ||} 
 \hline
  $T_x$ & 200 (Amazon Reviews), 30 (Tweets)  \\ 
 \hline
Deep Learning Library & Keras \\ 
 \hline
 Optimizer & Adam [$lr=0.005$, $beta_1=0.9$, $beta_2=0.999$, $decay=0.01$] \\ 
 \hline
 Maximum Epoch & 50 \\ 
 \hline
    Dropout & 0.4 \\ 
 \hline
 Early Stopping Patience & 3  \\ 
 \hline
 Batch Size & 32  \\ 
 \hline
 Validation Split & 0.15  \\ 
 \hline
 \end{tabular}
\end{center}
\end{table}
\fi

\subsection{COVID-19 Tweet Dataset}
For the COVID-19 use-case, we use Twitter posts collected using CitizenHelper~\cite{pandey2018citizenhelper} system in March 2020, for the geo-bounding box of the Washington D.C. Metro region. These tweets were annotated by volunteers of regional Community Emergency Response Teams (CERTs), with `\textit{Relevant}' label denoting how relevant a tweet is %such that it can useful 
for crisis response operations. The label values range on a scale %with values
of $1$-$4$. %For experiments, 
We convert them into binary classes % classification problem 
by considering values $1$ and $2$ as $-$ve ($0$) class and values $3$ and $4$ as $+$ve ($1$) class. This dataset consists of $4911$ tweets with $-$ve ($Relevant$=$0$) and $637$ tweets with $+$ve ($Relevant$=$1$) classes. Following unsupervised domain adaptation criteria, the filtering models are trained using only the TREC dataset and evaluated on the COVID-19 tweets. For each independent run of the experiment, a balanced subset of size $637$ for both classes is selected for testing.

\begin{table}
\scriptsize
%\footnotesize
\begin{center}
%\begin{tabular}{|c||p{0.5cm}|p{0.5cm}|p{0.6cm}|p{0.7cm}|p{0.6cm}|p{0.7cm}||p{0.8cm}| }
\begin{tabular}{|p{0.75cm}||p{0.45cm}|p{0.45cm}|p{0.6cm}|p{0.75cm}|p{0.6cm}|p{0.7cm}||p{0.8cm}| }
%\begin{tabular}{|c||p{0.5cm}|c|c|c|c|c|p{0.85cm}| }
    \hline
S $\rightarrow$ T  & {\color{blue}LR} & {\color{blue}SVM} & {\color{blue}CNN} & {\color{blue}BiLSTM} & AMN & HATN & {\color{blue}ST}\\
    \hline
B $\rightarrow$ K & 76.40 & 75.95  & 81.20 & 84.45 & 81.88 & 87.03 & 87.22 \\
B $\rightarrow$ E & 75.53 & 74.05  & 80.44 & 84.61 & 80.55 & 85.75 & 85.51 \\
B $\rightarrow$ D & 81.08 & 81.43  & 82.94 & 83.52 & 85.62 & 87.07 & 86.32 \\
K $\rightarrow$ B & 76.12 & 75.78  & 78.78 & 80.67 & 79.05 & 84.88 & 81.85 \\
K $\rightarrow$ E & 80.37 & 81.20  & 85.17 & 87.37 & 86.68 & 89.00 & 87.09 \\
K $\rightarrow$ D & 73.32 & 74.98  & 76.41 & 78.49 & 79.50 & 84.72 & 81.13 \\
E $\rightarrow$ B & 74.85 & 74.18  & 78.08 & 81.18 & 77.52 & 84.03 & 81.50 \\
E $\rightarrow$ K & 81.85 & 81.85 & 86.59 & 89.00 & 87.83 & 90.08 & 89.21 \\
E $\rightarrow$ D & 75.82 & 75.83 & 78.35 & 78.46 & 85.03 & 84.32 & 81.37 \\
D $\rightarrow$ B & 81.17 & 82.20 & 82.26 & 84.83 & 84.53 & 87.78 & 87.02 \\
D $\rightarrow$ K & 76.42 & 77.58  & 81.09 & 85.21 & 81.67 & 87.47 & 86.37 \\
D $\rightarrow$ E & 72.47 & 73.68  & 79.56 & 83.66 & 80.42 & 86.32 & 85.63 \\
\hline
AVG & 77.12 & 77.39 & 80.91 & 83.45 & 82.52 & 86.54 & 85.02 \\
    \hline
\end{tabular}
\end{center}
 \caption{Performance comparison (accuracy) of various models on the standard benchmark dataset of amazon reviews. Methods in {\color{blue} blue} do not use any unlabeled target data; hence relevant in our context. Each reported score is an average of 10 independent runs of each experiment.} \label{table:benchmark}
\end{table}

\begin{table}
\scriptsize
%\footnotesize
\begin{center}
\begin{tabular}{|c||c|c||c|c|p{0.85cm}| }
    \hline
Target  & LR & SVM & CNN & BiLSTM & ST\\
    \hline
Guatemala Earthquake & 60.14 & 56.76 & 60.47 & 65.54 & 59.97\\
Typhoon Yolanda   & 65.39 & 65.97 & 63.05 & 65.49 & 65.53\\
Australia Bushfire  & 65.61 & 63.23 & 62.10 & 60.10 & 62.44\\
Boston Bombings  & 71.47 & 75.45 & 69.72 & 71.43 & 72.08\\
Queensland Floods  & 65.56 & 64.81 & 64.13 & 66.01 & 66.21\\
Chile Earthquake  & 43.09 & 37.94 & 43.37 & 35.45 & 39.23\\ 
Typhoon Hagupit  & 49.86 & 46.22 & 49.21 & 54.13 & 52.61\\
Nepal Earthquake  & 57.11 & 55.39 & 58.61 & 60.49 & 61.35\\
Paris Attacks  & 71.43 & 71.72 & 72.50 & 72.14 & 71.31\\
Florida School Shooting  & 58.79 & 63.02 & 58.82 & 59.71 & 60.55\\
\hline
AVG  & 60.85 & 60.05 & 60.20 & 61.05 & 61.13\\
    \hline
\end{tabular}
\end{center}
 \caption{Performance comparison (accuracy) of unsupervised models on TREC-Priority (tweet) dataset showing that deep models are \textbf{\underline{not}} strictly superior than simpler models due to data sparsity. Each reported score is an average of 10 independent runs of each experiment. $Source$ = $Everything$ - $Target$.} \label{table:simple_benchmark}
\end{table}

\begin{table*}
\scriptsize
%\footnotesize
\begin{center}
\begin{tabular}{|c||*{2}{c}|*{2}{c}|*{2}{c}||*{2}{c}|*{2}{c}|*{2}{c}| }
\hline
TARGET    & \multicolumn{6}{c||}{\textbf{Priority}}
                & \multicolumn{6}{c|}{\textbf{Factoid}}                \\
    & \multicolumn{2}{c|}{ST}
            & \multicolumn{2}{c|}{ST-DAAN}
                    &
             \multicolumn{2}{c||}{MT-DAAN}
                    & \multicolumn{2}{c|}{ST}
                            & \multicolumn{2}{c|}{ST-DAAN}
                            & \multicolumn{2}{c|}{MT-DAAN}                \\
  &   Acc  &   F1  & Acc  &   F1  &    Acc  &   F1  &   Acc  &   F1  &   Acc  &   F1 & Acc  &   F1  \\
    \hline
Guatemala Earthquake & 59.97 & 62.39 & 69.07 & 69.66 & 69.05 & 69.34 & 68.92 & 68.47 & 79.90 & 80.76 & 84.05 & 97.01\\
Typhoon Yolanda & 65.53 & 65.47 & 66.07 & 63.73 & 67.42 & 67.30 & 80.50 & 84.42 & 82.71 & 85.61 &  84.36 & 86.93\\
Australia Bushfire & 62.44 & 66.69 & 61.07 & 63.42 & 61.93 & 64.28 & 64.58 & 60.69 & 65.64 & 60.53 & 65.04 & 60.13\\
Boston Bombings & 72.08 & 74.29 & 72.34 & 73.37 & 73.80 & 74.74 & 83.10 & 88.51 & 81.42 & 85.90  & 85.82 & 88.82\\
Queensland Floods & 66.21 & 65.94 & 67.19 & 66.97 & 66.74 & 66.46 & 37.56 & 48.90 & 50.46 & 59.82 & 49.52 & 59.21\\
Chile Earthquake & 39.23 & 40.92 & 38.91 & 42.37 & 41.80 & 46.33 & 30.38 & 33.97  & 39.87 & 48.68 & 45.28 & 54.58\\
Typhoon Hagupit & 52.61 & 50.59 & 58.97 & 58.94 & 57.50 & 57.52 & 68.98 & 70.79  & 71.42 & 72.44 & 69.49 & 70.08\\
Nepal Earthquake & 61.35 & 59.44 & 60.18 & 57.80 & 61.65 & 59.49 & 74.04 & 76.08  & 80.72 & 81.00 & 81.04 & 81.02\\
Paris Attacks & 71.31 & 76.26 & 70.42 & 74.08 & 74.44 & 77.21 & 75.78 & 80.35  & 82.35 & 84.89 & 82.52 & 85.63\\
Florida School Shooting & 60.55 & 61.75 & 65.47 & 64.07 & 62.51 & 63.24 & 76.73 & 82.67  & 84.55 & 87.51 & 85.80 & 88.15\\
\hline
AVG & 61.13 & 62.37 & 62.97 & 63.44 & \textbf{63.68} & \textbf{64.59} & 66.06 & 69.49  & 71.90 & 74.71 & \textbf{73.29} & \textbf{77.16}\\
    \hline
\end{tabular}

\begin{tabular}{|c||*{2}{c}|*{2}{c}|*{2}{c}||*{2}{c}|*{2}{c}|*{2}{c}|}
\hline
TARGET    & \multicolumn{6}{c||}{\textbf{Sentiment}}
                & \multicolumn{6}{c|}{\textbf{Irrelevant}}                \\
    & \multicolumn{2}{c|}{ST}
    & \multicolumn{2}{c|}{ST-DAAN}
            & \multicolumn{2}{c||}{MT-DAAN}
                    & \multicolumn{2}{c|}{ST}
                    & \multicolumn{2}{c|}{ST-DAAN}
                            & \multicolumn{2}{c|}{MT-DAAN}                \\
  &   Acc  &   F1  &   Acc &   F1 &   Acc  &   F1 &   Acc  &   F1 &   Acc  &   F1  &   Acc  &   F1  \\
    \hline
Guatemala Earthquake & 96.96 & 97.03 & 96.45 & 96.68 & 96.76 & 92.73 & 89.36 & 89.03 & 91.22 & 91.06 &  93.11 & 92.73\\
Typhoon Yolanda & 75.81 & 77.62 & 77.54 & 79.01 &  76.82 & 78.35 & 76.05 & 79.77 & 78.49 & 80.59 &  80.46 & 82.31\\
Australia Bushfire & 75.95 & 77.58 & 78.80 & 79.12 &  78.54 & 78.92 & 35.42 & 47.164 & 53.78 & 65.11 &  51.76 & 63.36\\
Boston Bombings & 81.39 & 81.11 & 80.73 & 80.70 &  82.13 & 82.10 & 58.15 & 55.73 & 58.15 & 57.43 &  61.49 & 61.45\\
Queensland Floods & 81.69 & 80.39 & 81.05 & 81.39 &  81.53 & 81.32 & 65.68 & 65.36  & 67.26 & 65.72 & 67.88 & 67.27\\
Chile Earthquake & 92.69 & 92.91 & 93.10 & 93.21 &  93.62 & 93.68 & 75.16 & 84.98  & 80.46  & 86.38 & 80.64 & 86.56\\
Typhoon Hagupit & 84.98 & 85.86 & 85.15 & 86.14 &  85.43 & 86.38 & 63.21 & 75.04  & 71.50 & 78.25 & 70.22 & 77.27\\
Nepal Earthquake & 67.75 & 68.42 & 70.20 & 70.51 &  69.96 & 70.31 & 31.79 & 42.10  & 36.97 & 47.41 & 41.49 & 52.87\\
Paris Attacks & 76.01 & 76.63 & 73.65 & 73.98 &  74.47 & 74.60 & 33.91 & 35.25  & 44.52  & 48.32 & 47.17 & 51.32\\
Florida School Shooting & 68.77 & 71.77 & 67.06 & 70.03 &  68.14 & 71.05 & 32.66 & 40.90  & 44.22 & 55.27 & 47.64 & 58.65\\
\hline
AVG & 80.20 & 80.93 & 80.37 & \textbf{81.08} &  \textbf{80.74} & 80.94 & 56.14 & 61.53  & 62.66 & 67.55 & \textbf{64.19} & \textbf{69.38}\\
    \hline
\end{tabular}
\end{center}
 \caption{Unsupervised domain adaptation results on TREC dataset showing performance boost for \textit{\textbf{Priority}}, \textit{\textbf{Factoid}}, and \textit{\textbf{Irrelevant}} tasks. However, \textit{\textbf{Sentiment}} task did not show a significant improvement. See performance evaluation section for details. Each reported score is an average of 10 independent runs of each experiment.} \label{table:trec_result}
\end{table*}

\begin{table}
\scriptsize
%\footnotesize
\begin{center}
\begin{tabular}{|c||*{2}{c}|*{2}{c}|*{2}{c}||}
\hline
TARGET    & \multicolumn{6}{c||}{\textbf{Relevant}}            \\
    & \multicolumn{2}{c|}{ST}
            & \multicolumn{2}{c|}{ST-DAAN}
                    &
             \multicolumn{2}{c||}{MT-DAAN}             \\
  &   Acc  &   F1  & Acc  &   F1  &    Acc  &   F1   \\
    \hline
COVID-19 & 73.25 & 77.36 & 74.55 & 77.51 & 77.00  & 78.09\\
    \hline
\end{tabular}
\end{center}
 \caption{Unsupervised domain adaptation results for COVID-19 tweets using only the TREC dataset for training. Each reported score is an average of 10 independent runs of each experiment.} \label{table:covid}
\end{table}

\section{Results \& Discussion}
%For all our experiments, reported scores are based on an average of 5 independent runs of each experiment.

%\subsection{Validation of Single-Task Architecture}

We first validate the performance of the adopted unsupervised ST model \cite{jitin} by comparing it with the following standard neural network architectures and state-of-the-art models used for domain adaption in text. We use the standard benchmark dataset of Amazon reviews. Following the traditional domain adaptation experimental setup, each experiment represented as S $\rightarrow$ T consists of a source domain (S) on which the model is trained and a target domain (T) on which the model is tested. We use Keras deep learning library for our implementations; with $T_{x}$=$200$ for Amazon reviews and $30$ for Tweets. We use Adam optimizer with a dropout of $0.4$, maximum epoch of $50$, early stopping patience of $3$, batch size of $32$, and validation split of $0.15$. %Implementation details are shown in \textbf{Table \ref{table:implementation}}.
\begin{enumerate}
    \item \textbf{Simple Baselines:} We construct simple baseline classifiers \cite{scikit-learn}: Logistic Regression (LR) and Support Vector Machines (SVM). The input to these models are constructed by aggregating the $300$-dimensional word embeddings of words in each review.
    \item \textbf{CNN:} A standard Convolutional Neural Network inspired by Kim, 2014  \cite{kim2014convolutional} is constructed with the following architecture:\\
    $ Word\ Embeddings (T_x,300) \rightarrow Conv1D (128, 5)$ \\ $\rightarrow  MaxPooling1D (5)$ $\rightarrow Conv1D (128, 5) \\ \rightarrow  MaxPooling1D (5)$ $ \rightarrow Conv1D (128, 5) \\ \rightarrow  GlobalMaxPooling1D() \rightarrow Dense (128)$ \\$ \rightarrow Dense (2) \rightarrow y$.\\
    This is combined with dropouts, \textit{relu} activations, and ending with \textit{softmax} activation producing labels for binary classification. State-of-the-art deep learning methods for existing social media mining approaches of crisis analytics \cite{alam2018domain,nguyen2016rapid} use a similar architecture.
    \item \textbf{BiLSTM:} This is the bottom-most layer in \textbf{Figure \ref{fig:mtl}} with the activation $a^{<T_x>}$ passed through the following: $Dense(10) \rightarrow Dense(2) \rightarrow y$ also including dropouts, \textit{relu} activation, and ending with \textit{softmax}.
    \item \textbf{AMN and HATN:} AMN \cite{amn} and HATN \cite{li2018hierarchical} are attention-based methods which use gradient reversal to perform domain adversarial training on the unlabeled data from source and target domains. HATN is an extension to AMN by adding the hierarchical component and jointly training pivot and non-pivot networks.
\end{enumerate}
Input to all the models are word vectors\footnote{\url{https://code.google.com/archive/p/word2vec/}} \cite{mikolov2013distributed}.
The evaluation on amazon reviews shows how well the single-task (ST) model perform when compared to the existing top-performing domain adaptation models on benchmark dataset. \textbf{Table \ref{table:benchmark}} shows accuracy scores on the Amazon cross-domain sentiment analysis dataset. HATN uses unlabeled target data, gradient reversal, explicit pivot extraction, and joint training making it a computationally expensive method. As shown in the experimental evaluation, we use the same Amazon dataset and GoogleNews word vectors for our experiments. ST, being \textit{\textbf{unsupervised}} with no need of unlabeled target data, performed competitively with an overall accuracy of 85.02\%; thus establishing a strong fully unsupervised building block for us to build upon.

\subsection{Crisis Tweets vs Amazon Reviews}

\textbf{Table \ref{table:benchmark}} and \textbf{\ref{table:simple_benchmark}} show that deep models struggle with small datasets such as TREC-IS tweets. When ST model outperformed Logistic Regression by $\sim8\%$ on the Amazon reviews dataset, the difference was only less than $1\%$ with no statistical significance on the TREC-Priority dataset. Note that we conduct experiments with various parameter combinations on the deep models when using tweets. For example, $T_x=200$ for amazon reviews and $T_x=30$ for tweets due to the difference in their average word-length.  \textit{Books} domain of Amazon reviews has $182$ average number of tokens per review with a vocab size of $105920$. On the other hand, the event with highest number of tweets in the TREC dataset (Paris Attacks) has only 18.62 average number of tokens per tweet with a vocab size of $4152$. This difference makes it intuitively challenging to train deep models with several parameters that may lead the model to memorize the entire dataset resulting in poor generalization. Multi-task learning and domain adversarial training try to alleviate this problem by training the shared BiLSTM layer with much more data from different tasks and unlabeled data.

\iffalse
\begin{table}[!htbp]
%\scriptsize
\footnotesize
\begin{center}
 \caption{Performance comparison (accuracy) of four relevant word embedding models on TREC-Sentiment task showing that the tweet-based embeddings such as Glove or CrisisNLP  did  not  significantly outperform other  models.} \label{table:wv}
\begin{tabular}{|c|p{0.7cm}|p{1.2cm}|p{0.5cm}|p{1.0cm}| }
    \hline
TARGET    & fastText \cite{mikolov2018advances}
            & GoogleNews \cite{mikolov2013distributed}
                    & Glove \cite{pennington2014glove}
                            & CrisisNLP \cite{imran2016lrec}                \\
    \hline
Guatemala Earthquake & 96.96 & 95.72 & 95.27 & 97.97\\
Typhoon Yolanda & 75.81 & 83.30 & 86.49 & 79.41\\
Australia Bushfire & 75.95 & 79.35  & 80.24  & 75.86 \\
Boston Bombings & 81.39 & 82.43 & 80.55  & 81.16 \\
Queensland Floods & 81.69  & 84.06  & 84.01  & 81.50 \\
Chile Earthquake & 92.69  & 92.82  & 92.93  & 92.60 \\
Typhoon Hagupit & 84.98 & 87.55  & 84.25 & 88.76\\
Nepal Earthquake & 67.75  & 68.46 & 73.63  & 67.50 \\
Paris Attacks & 76.01  & 77.67  & 74.49  & 77.43 \\
Florida School Shooting & 68.77  & 66.84  & 66.97  & 65.06\\
 \hline
AVG  &   80.20   &   81.82   &   81.88  &   80.73 \\
    \hline
\end{tabular}
\end{center}
\end{table}
\fi

\subsection{MT-DAAN Performance Evaluation}
The primary purpose of the MT-DAAN model is to show that sharing the bottom layer of the model (i.e., shared representation) for different tasks along with domain adversarial training can help improve the generalizability of some of the tasks that are otherwise trained alone in the single-task model. The experiments for MT-DAAN are setup in the same unsupervised way as for single-task. No data from the test crisis is used for training. For example, if we are testing our model for the event `\textit{Typhoon Yolanda}', no data from this crisis is used for training. Note that the domain classifier component uses unlabeled data only from rest of the crisis; making it a fully unsupervised domain adaptation approach. Performance scores of the four tasks (\textit{Priority}, \textit{Factoid}, \textit{Sentiment}, and \textit{Irrelevant}) are shown in \textbf{Table \ref{table:trec_result}}. The results show clear performance improvement for \textit{Priority}, \textit{Factoid}, and \textit{Irrelevant} tasks. However, \textit{Sentiment} task did not show significant improvement. We speculate that this is because other tasks do not generalize the bottom layer enough to boost the sentiment classification performance. These results show the usefulness of multi-task learning as well as domain adversarial training where different tasks in multiple domains help each other when the data is sparse and labels are limited. 

%\subsection{MT-DAAN Hyperparameters}

%Apart from the generic hyperparameters mentioned in \textbf{Table \ref{table:implementation}}, multi-task learning has specific hyperparameters that can be tuned to improve performance. The domain adversarial training hyperparameters, $w_d$ and $\lambda$, are set to $0.1$ and $0.4$ respectively (similar to \textit{ST-DAAN}). $w_d$ is the domain adversarial loss weight and $\lambda$ is the strength of the reversal (refer section 3.3 for more details). Furthermore, the loss weights ($w_p$, $w_f$, $w_s$, and $w_i$) corresponding to the 4 tasks (\textit{Priority}, \textit{Factoid}, \textit{Sentiment}, and \textit{Irrelevant}), can be set differently to give prominence to the task under consideration. A baseline setting is to set all weights to $1.0$. However, this forces the model to focus on all tasks equally. To provide more flexibility for the model to learn specific tasks, the weights can be changed accordingly. For example, [$w_p$=$1.0$, $w_f$=$0.3$, $w_s$=$0.1$, and $w_i$=$0.6$] is a \textit{Priority} task which gives more prominence to \textit{Irrelevant} class label and less prominence to \textit{Sentiment} class label. We perform a simple grid search for values in range [$0.1$-$1.0$] with $0.1$ interval to find corresponding weight combination for each task.

\begin{figure}
  \centering
  \includegraphics[width=6.5cm]{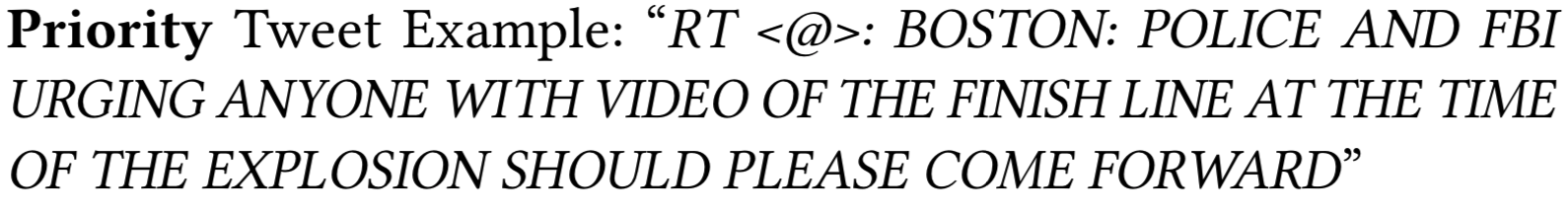}
  \includegraphics[width=7cm]{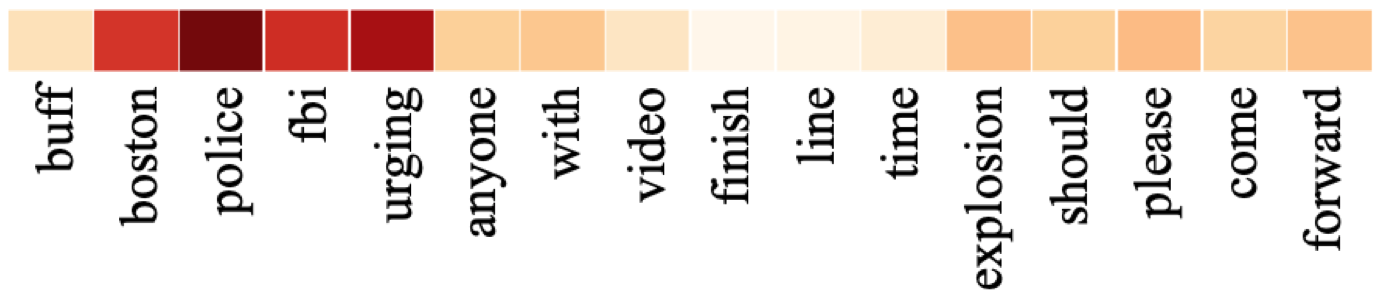}
  \includegraphics[width=6.5cm]{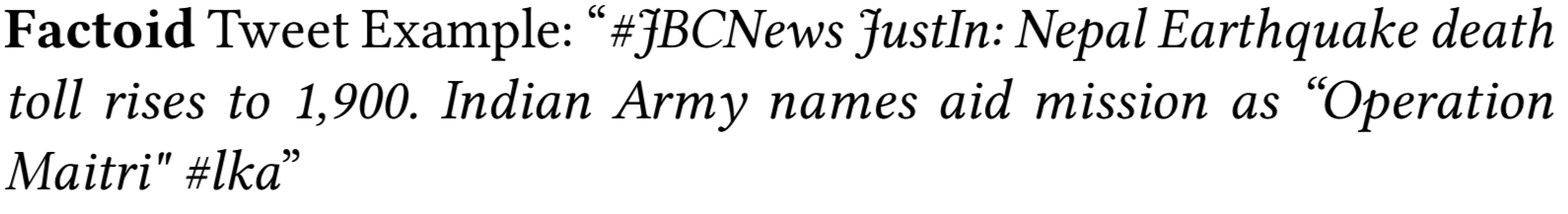}
  \includegraphics[width=6.8cm]{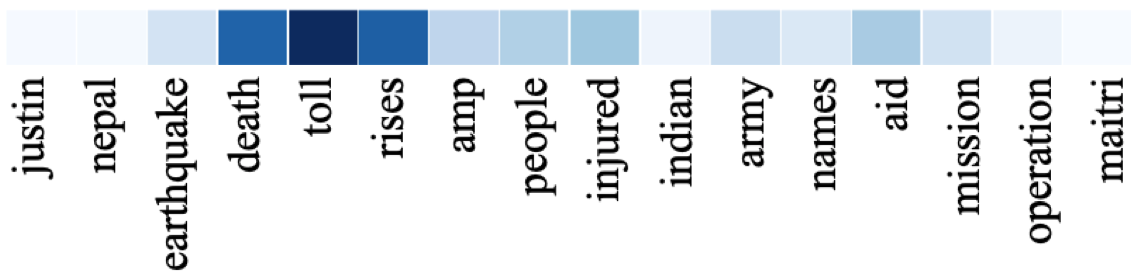}
  \includegraphics[width=6.5cm]{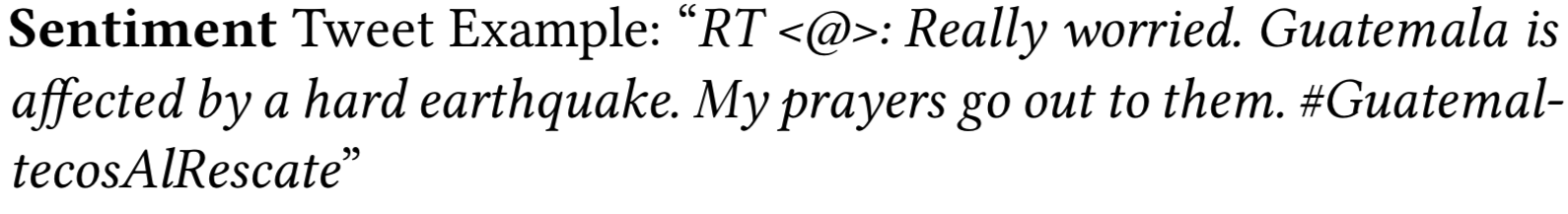}
  \includegraphics[width=3cm]{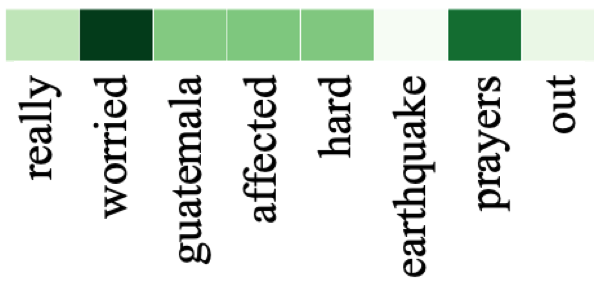}
  \includegraphics[width=6.5cm]{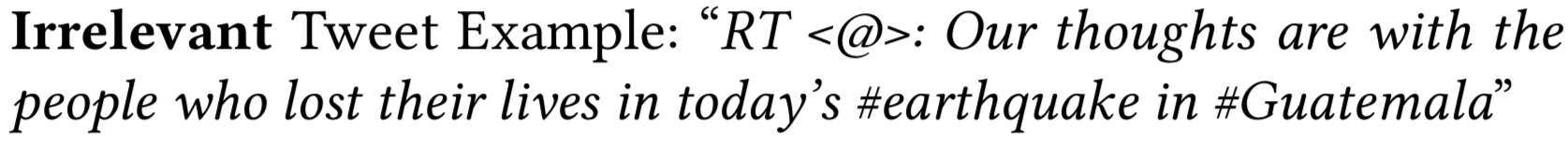}
  \includegraphics[width=3cm]{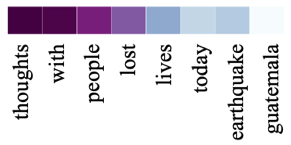}
 \caption{Examples of interpretable results using attention; darker the shade, higher the attention. Recall that no data from the crisis-event for testing is used for training the model. Even then, relevant keywords such as \textit{`police urging'}, \textit{`death toll rises'}, \textit{`worried'}, and \textit{`thoughts with people'} are correctly picked up by the attention layers of their respective tasks. }  \label{fig:attn_viz}
\end{figure}

\subsection{Word Vectors} 
We use fastText\cite{mikolov2018advances} as our word embeddings for tweets because of its sub-word usage and the ability to create vectors for arbitrary and out-of-vocabulary words. Although there exists many alternatives, picking the one that works well for a specific dataset is not trivial. We conducted experiments using four choices of word embeddings: \textit{fastText} \cite{mikolov2018advances}, \textit{GoogleNews} \cite{mikolov2013distributed}, \textit{Glove} \cite{pennington2014glove}, and \textit{CrisisNLP} \cite{imran2016lrec}. Averaging over 10 crises, we received the following accuracy scores (in \%) respectively for the above word embeddings: \{$80.20$, $81.82$, $81.88$, $80.73$\}. Unlike fastText, we fine-tune these pre-trained vectors %using Gensim \cite{gensim} 
to create vectors for out-of-vocabulary words. Vectors for words that are already in the vocabulary are \textit{locked} while tuning for consistency in evaluation. The tweet-based embeddings such as Glove or CrisisNLP did not significantly outperform other models. Glove vectors are 200-dimensional while the rest are 300-dimensional which makes the experiment favoring Glove word vectors. This experiment shows that the problem of finding a strictly superior word vector model for tweets still remains a challenging task.

\begin{figure}
  \centering
  \includegraphics[width=8.6cm]{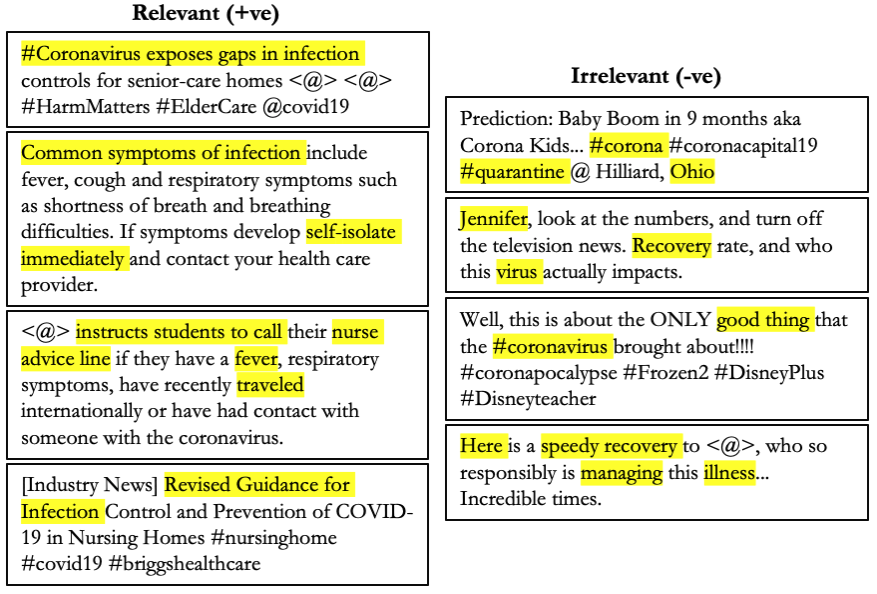}
 \caption{Examples of interpretable results using attention for relevancy prediction of COVID-19 tweets. With $77\%$ accuracy, although the highly attended words in the `Relevant' tweets provide some intuitive sense of interpretability, the highlighted words in the `Irrelevant' tweets are somewhat ambiguous because it is unclear if those words are chosen due to their specific or generic nature. This shows both the benefits and challenges of unsupervised and interpretable domain adaptation.}  \label{fig:attn_viz_covid}
\end{figure}

\subsection{Interpretability: Attention Visualization}
The attention weights used to create the context vector by the dot product operation with word activations represent the interpretable layer in our architecture. These weights represent the importance of each word in the classification process. Some examples are shown in \textbf{Figures \ref{fig:attn_viz}} and \textbf{\ref{fig:attn_viz_covid}}. Stronger the color intensity stronger the word attention. In the first example, `\textit{boston police urging}' is the reason why the tweet is classified as $+$ve priority. Similarly, `\textit{death toll rises}' in the \textit{Factoid} example, `\textit{worried, prayers}' in the \textit{Sentiment} example, and `\textit{thoughts with people}' in the \textit{Irrelevant} example are clear intuitive indicators of +ve predictions. These examples show the importance of having interpretability as a key criterion in crisis domain adaptation tasks for social media.

To the best of our knowledge, in social media mining for crisis analytics, there does not exist a ground truth dataset that highlights the words that explain the labels for tweets. Using our model as a guide, we hope to build a robust evaluation dataset as our immediate next step so that the models can be quantitatively evaluated using robust trust-evaluation methods such as LIME \cite{ribeiro2016should}. It is also crucial to note that binary classification tasks such as sentiment analysis of Amazon reviews has a clear class divide that produces intuitive keywords such as `good', `excellent', or `great' for $+$ve reviews and `bad', `poor', or `horrible' for $-$ve reviews. However, for short texts such as tweets shown in \textbf{Figure \ref{fig:attn_viz_covid}}, `relevancy' can depend on the context and it is unclear which keywords truly represent the examples in the `irrelevant' class.

\section{COVID-19 Use-Case}
We show a practical implication of our work by applying it to COVID-19 tweets described in Section 4.3. Our goal is to interpretably predict if a COVID-19 tweet is relevant or not; a binary classification task. The models are trained using only the TREC dataset and evaluated on the COVID-19 tweets (a balanced subset of size $637$ for $+$ve and $-$ve labels). We found that a combination of `\textit{Priority}' and `\textit{Irrelevant}' labels from TREC performs better to predict COVID-19's `\textit{Relevant}' label (this can be trivially verified by constructing two binary classifiers). We augment all three methods (\textit{ST}, \textit{ST-DAAN}, and \textit{MT-DAAN}) with an additional condition before label prediction: $R_{c} = P_{t} \cap \overline{I_{t}}$, which means that a COVID-19 tweet is `\textit{Relevant}' only if it is predicted both `\textit{Priority}' = $1$ and `\textit{Irrelevant}' = $0$. The scores are reported in \textbf{Table \ref{table:covid}} and the attention results are shown in \textbf{Figure \ref{fig:attn_viz_covid}}, demonstrating the effectiveness of our proposed method.

%\section{Future Work}
%Empirical evaluation of word vector models showed that for practical purposes with challenging datasets, tweet-trained models such as Glove or CrisisNLP did not significantly outperform other models such as fastText or GoogleNews vectors. This calls for further investigation to improve word embeddings for tweets, specifically in low resource setting. Another direction is to address the non-transferable components. For example, \textit{Sentiment} is easier to detect than \textit{Factoid} because there is more linguistic overlap in how sentiments are expressed in different domains. However, Factoid phrases such as `\textit{Indian army names aid mission}' shown in \textbf{Figure \ref{fig:attn_viz}} are more challenging to find transferable representations. This calls for methods such as a combination of named entity recognition and transfer learning in crisis analytics; which is not currently attempted. Furthermore, our work addresses only binary class labels. A useful next direction would be to extract more fine-grained  information; for example, classifying tweets into a discrete set of priority labels such as $1$-$4$ or ranking them.

\section{Conclusion}

We presented a novel approach of unsupervised domain adaptation with multi-task learning to classify relevant information from Twitter streams for crisis management, while addressing the problems of data sparsity and limited labels. We showed that a multi-task learning model that shares the lower layers of the neural network with dedicated attention layers for each task along with a domain classifier branch can help improve generalizability and performance of deep models in the settings of limited data. % low resource settings. 
Furthermore, we showed that using an attention-based architecture can help in interpreting the classifier's predictions by highlighting the important words that justify the predictions. We also presented an in-depth empirical analysis of the state-of-the-art models on both benchmark dataset of Amazon reviews and TREC dataset of crisis events. 
The application of our generic approach for interpretable and unsupervised domain adaptation within a multi-task learning framework %for social web data filtering 
can benefit social media mining systems in diverse domains beyond crisis management. \\
\noindent \textbf{Reproducibility:} Source code and instructions for deployment are available at  - {\color{blue}\url{https://github.com/jitinkrishnan/Crisis-Tweet-Multi-Task-DA}}.

\bibliographystyle{IEEEtran}
\bibliography{acmart}

\end{document}